\documentclass[sigplan]{acmart}[9pt]
\AtBeginDocument{%
  \providecommand\BibTeX{{%
    \normalfont B\kern-0.5em{\scshape i\kern-0.25em b}\kern-0.8em\TeX}}}

\usepackage{url}
\usepackage{enumitem}

\usepackage[utf8]{inputenc} %
\usepackage[T1]{fontenc}    %
\usepackage{hyperref}       %
\usepackage{url}            %
\usepackage{booktabs}       %
\usepackage{amsfonts}       %
\usepackage{nicefrac}       %
\usepackage{microtype}      %
\usepackage{xcolor}         %
\usepackage{colortbl}
\usepackage{subcaption}
\usepackage{graphicx}
\usepackage{svg}
\usepackage{multirow}
\usepackage{wrapfig}
\usepackage{threeparttable}
\usepackage{tikz}
\usepackage{makecell}
\usepackage{tablefootnote}
\usepackage{pifont}%
\usepackage{cellspace}
\usepackage{makecell}

\usepackage{color}
\newcommand*\circled[1]{\raisebox{.4pt}
                    {\tikz[baseline=(char.base)]{
            \node[shape=circle,draw,inner sep=1pt, style={fill=black, text=white}, scale=0.75] (char) {\textbf{#1}};}}}

\usepackage{tikz}
 
\newcommand*\halfcirc[1][1ex]{%
  \begin{tikzpicture}
  \draw[fill] (0,0)-- (90:#1) arc (90:270:#1) -- cycle ;
  \draw (0,0) circle (#1);
  \end{tikzpicture}}

\usepackage{pgfplots}
\pgfplotsset{compat=1.17}

\newcommand{\tabfpgasotacompare}{%
\begin{table}
    \centering
    \footnotesize
    \begin{tabular}{p{0.047\textwidth}|p{0.19\textwidth}|p{0.18\textwidth}}
    \toprule
\textbf{Feature}    & \textbf{SOTA FPGA Accelerators} & \textbf{Ours (GenGNN)} \\
\midrule

\textbf{Generic} & \ding{55}: GCN only~\cite{zhang2020hardware, zhang2021boostgcn}; \ding{55}: heavily rely on the property of a specific graph (e.g., sparsity)~\cite{zhang2021boostgcn} & \ding{52}: Support a wide types of GNNs; \ding{52}: GNN, dataset and graph structure agnostic\\ 

\hline 

\textbf{Real-time} & \ding{55}: Requires heavy graph preprocessing~\cite{zhang2020hardware} or partitioning on CPU~\cite{zhang2021boostgcn}  & \ding{52}: Zero preprocessing; directly takes in raw graphs and processes on FPGA \\

\hline
\textbf{Open-source} & \ding{55}: Not yet & \ding{52}: Open-source with executables upon publication \\

\bottomrule
    
    \end{tabular}
    \caption{Comparisons between our \textbf{GenGNN} framework and existing FPGA-based accelerators; our main advantages are \textit{generic}, \textit{real-time}, and \textit{open-source}.}
    \label{tab:fpga-sota-compare}
    \vspace{-10pt}
\end{table}}

\newcommand{\tabgnnlist}{%
\begin{table}
    \centering
    \footnotesize
    \begin{tabular}{@{\hskip 0.003\textwidth}p{0.038\textwidth}| p{0.04\textwidth} | p{0.028\textwidth} | p{0.281\textwidth}@{}}
    \toprule
    \textbf{Model} & \textbf{SOTA} & \textbf{Ours} & \textbf{Representativeness} \\
    \midrule
    \textbf{GCN} \cite{kipf2016semi}     & \ding{52} \textbf{F}$^{*}$~\cite{zhang2020hardware} & \ding{52} & GNN family that can be represented as sparse matrix-matrix multiplications (SpMM) \\
    
    \hline    
    
    \textbf{GIN} \cite{xu2018powerful}    & {\halfcirc} \textbf{A}$^{\text{\ding{61}}}$~\cite{yan2020hygcn} & \ding{52} & GNN family with edge embedding and transformation where SpMM \textit{does not} apply  \\
    
    \hline
    
    \textbf{GAT} \cite{velivckovic2017graph} & {\halfcirc} \textbf{A}$^{\text{\ding{61}}}$~\cite{auten2020hardware} & \ding{52} & GNN family with self-attention and possibly edge embeddings\\
    
    \hline
    
    \textbf{PNA} \cite{corso2020principal} & \ding{55} &  \ding{52} & A popular GNN family arbitrarily using multiple aggregation methods \\
    
    \hline
    
    \textbf{DGN} \cite{beani2021directional} & \ding{55} & \ding{52} & A state-of-the-art GNN with a directional flow at each node and  guided aggregation\\
    
    \hline
    
    \textbf{VN} \cite{gilmer2017neural} & \ding{55} & \ding{52} & A widely used GNN technique with a virtual node connecting to all other nodes\\
    \midrule
    
    \multicolumn{4}{p{0.45\textwidth}}{\textbf{F$^{*}$:} FPGA implementation; \textbf{A$^{\text{\ding{61}}}$:} ASIC implementation; \textbf{GCN:} graph convolutional network; \textbf{GIN:} graph isomorphism network; \textbf{GAT:} graph attention network; \textbf{PNA:} principal neighbourhood aggregation; \textbf{DGN:} directional graph network; \textbf{VN:} GNN with virtual node} \\
    
    \bottomrule
    
    \end{tabular}
    \caption{Currently prototyped representative GNNs by our framework \textbf{GenGNN} with easy extension to more types. }
    \label{tab:GNN-list}
    \vspace{-10pt}
\end{table}}

\newcommand{\figsparseformats}{%
\begin{figure}
    \centering
    \includegraphics[width=0.44\textwidth]{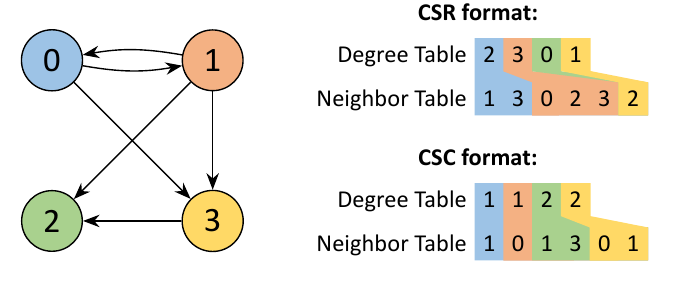}
    \vspace{-8pt}
    \caption{An example graph and the equivalent representations of its adjacency matrix in CSR and CSC formats.}
    \label{fig:sparse_formats}
\end{figure}}

\newcommand{\figarchoverall}{%
\begin{figure*}
    \centering
    \includegraphics[width=0.97\textwidth]{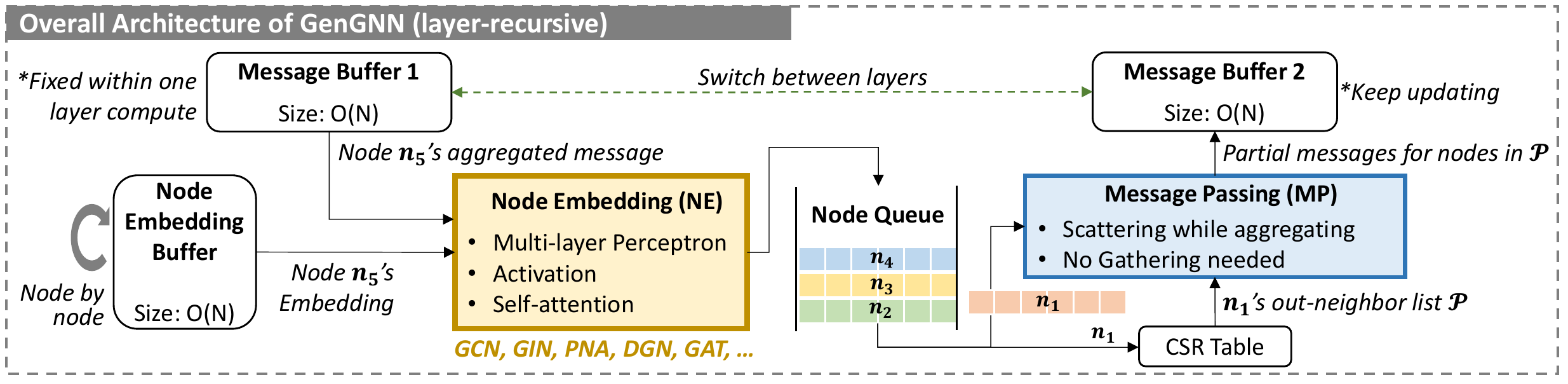}
    \vspace{-12pt}
    \caption{GenGNN overall architecture. It follows message passing mechanism and thus accommodates a wide range of GNNs.}
    \label{fig:arch-overall}
\end{figure*}}

\newcommand{\figpipeliningillustration}{%
\begin{figure}
    \centering
    \includegraphics[width=0.44\textwidth]{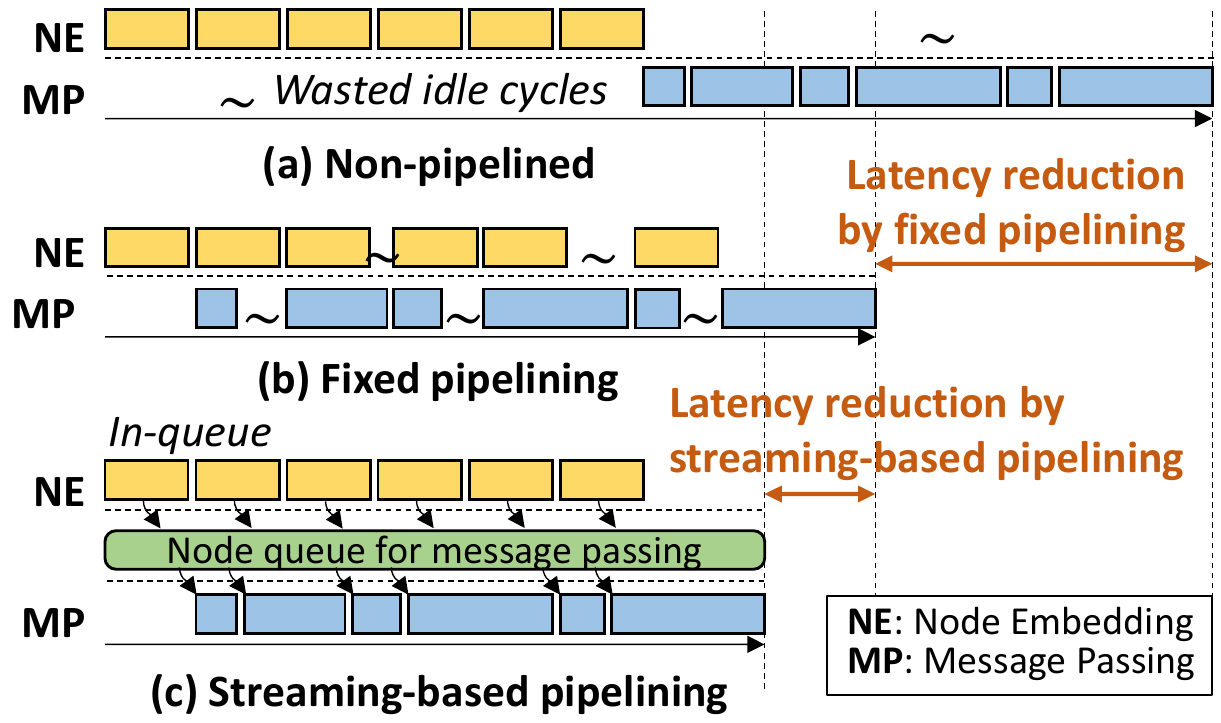}
    \caption{Different strategies of pipelining for node embedding and message passing.}
    \label{fig:pipelining-illustration}
\end{figure}}

\newcommand{\figgnnoverall}{%
\begin{figure*}
    \centering
    \includegraphics[width=0.95\textwidth]{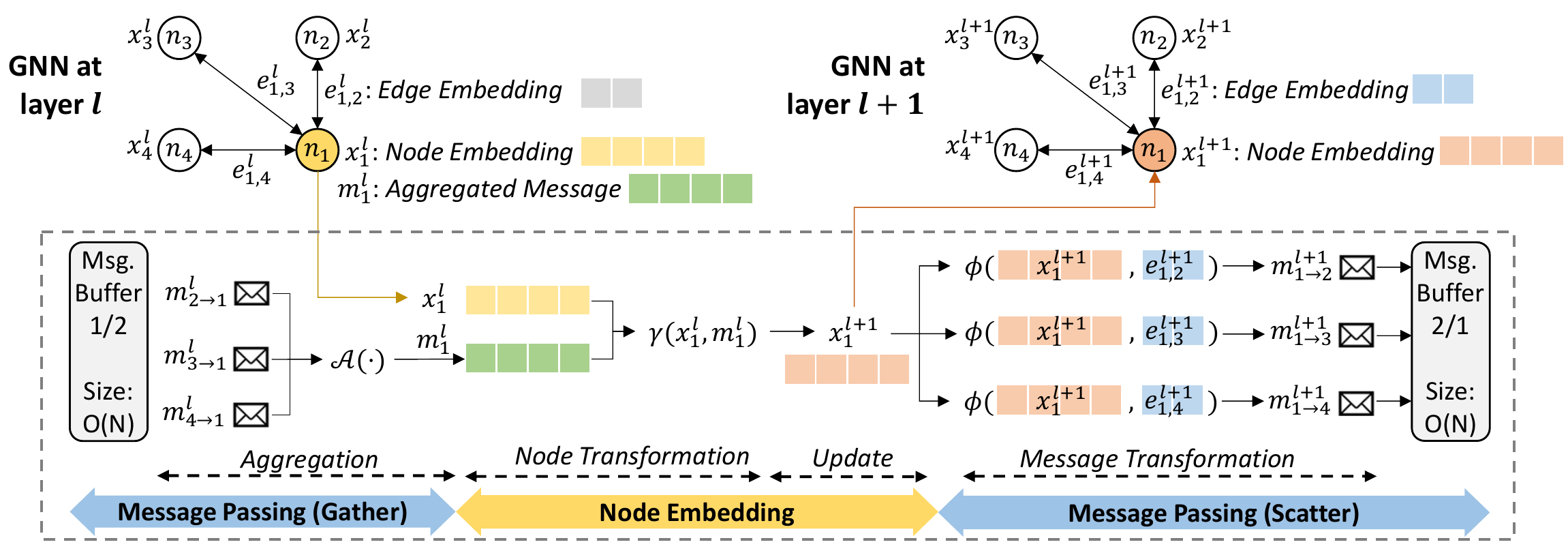}
    \vspace{-12pt}
    \caption{The generic message passing computation flow of prevailing GNN models.}
    \label{fig:gnn-overall}
\end{figure*}}

\newcommand{\figvirtualnode}{%
\begin{figure}
    \centering
    \includegraphics[width=0.47\textwidth]{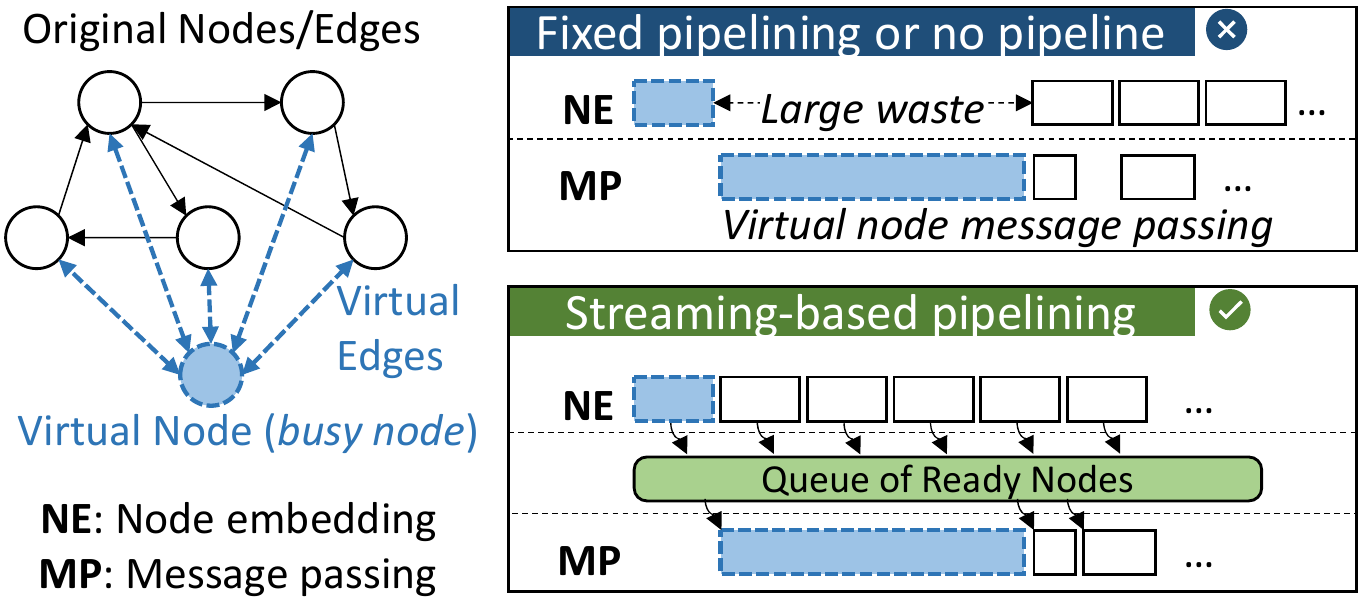}
    \caption{Streaming-based approach is especially beneficial for models with virtual nodes.}
    \label{fig:virtual-node}
\end{figure}}

\newcommand{\figlatencystats}{%
\begin{figure}
    \centering
    \includegraphics[width=0.99\linewidth]{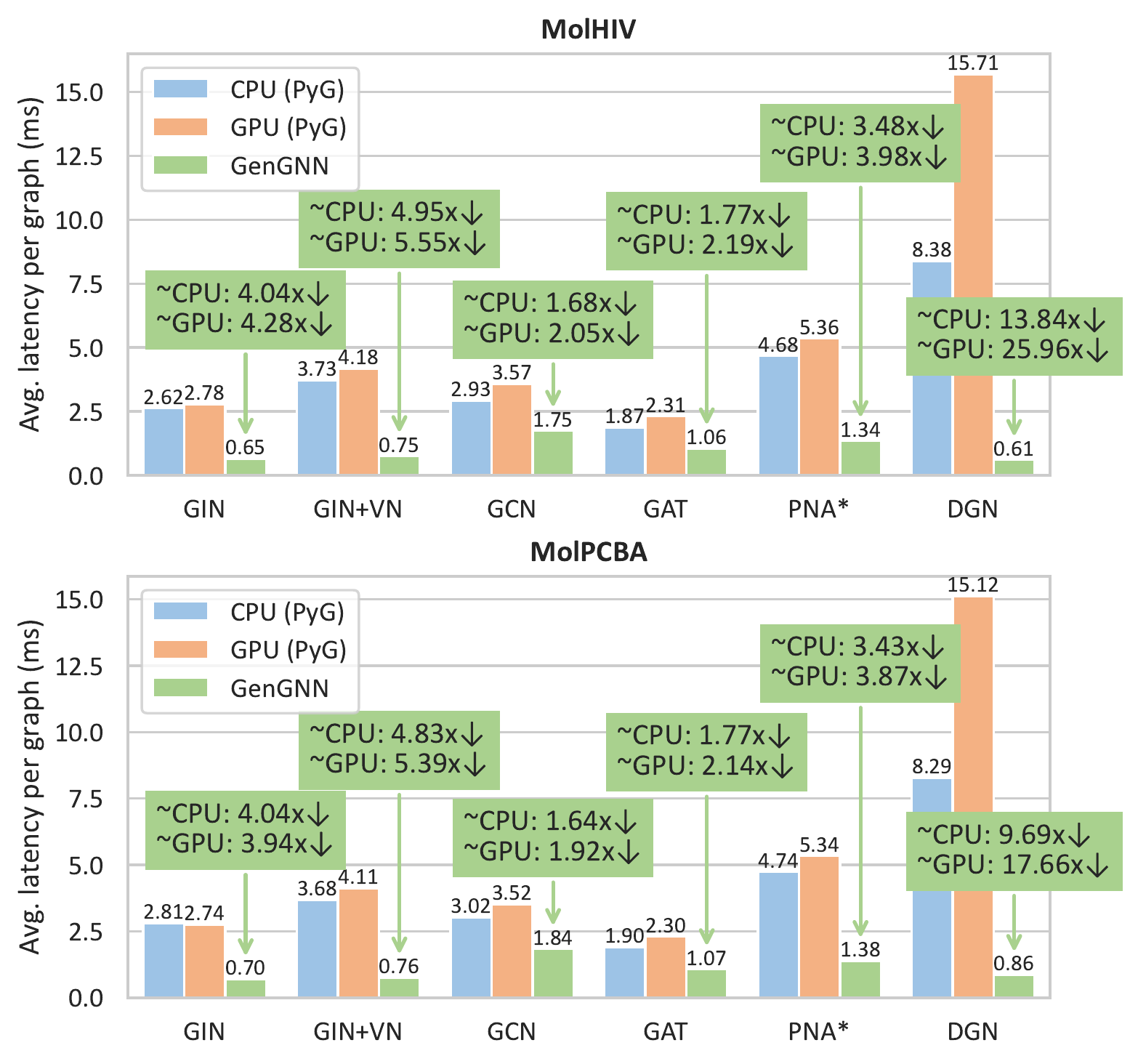}
    \vspace{-8pt}
    \caption{Average latency over test graphs. Top shows MolHIV results; bottom shows MolPCBA results. Results for the PNA* model are currently estimates from the Vitis HLS tool.}
    \label{fig:latency_stats}
\end{figure}}

\newcommand{\figlargelatencystats}{%
\begin{figure}
    \centering
    \includegraphics[width=0.95\linewidth]{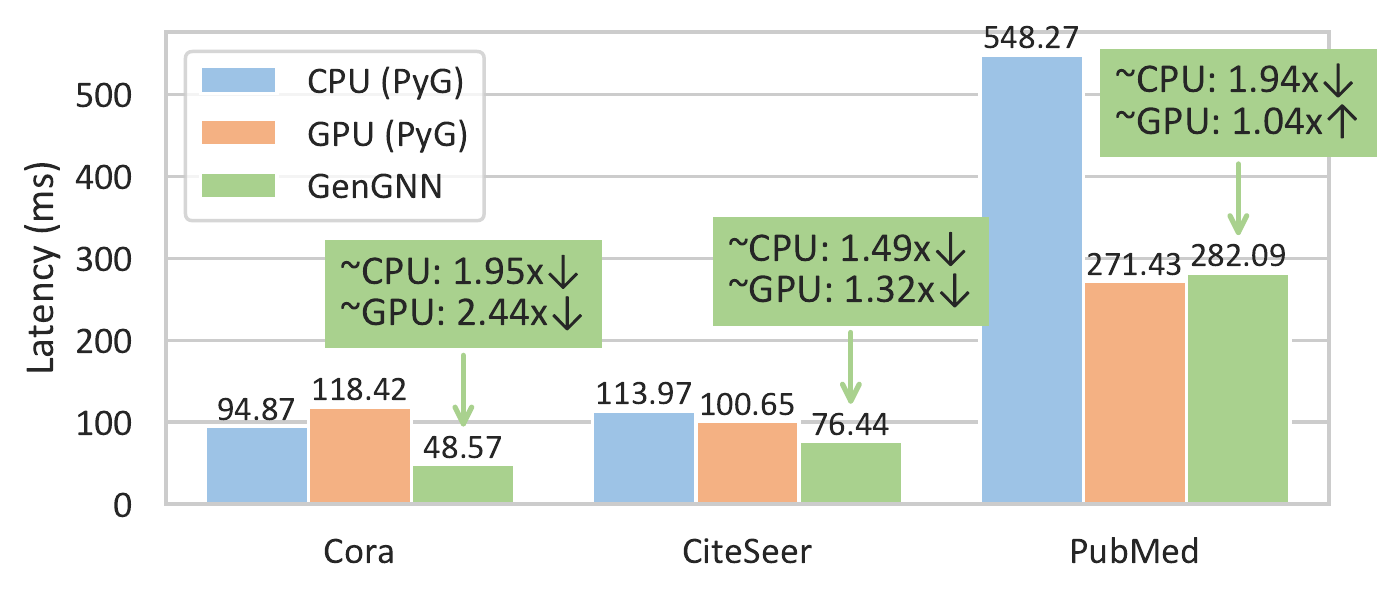}
    \vspace{-8pt}
    \caption{Latency of GenGNN DGN with Large Graph Extension over Cora, CiteSeer, and PubMed graphs.}
    \label{fig:lge_latency_stats}
\end{figure}}

\newcommand{\tabresource}{%
\begin{table}
    \centering
    \footnotesize
    \begin{tabular}{c|c|c|c|c|c}
    \toprule
    \textbf{Model}     &  \textbf{DSP} & \textbf{LUT} & \textbf{FF} & \textbf{BRAM} & \textbf{URAM} \\ 
    \midrule
    \textbf{Available} & 5,952 & 872K & 1,743K & 1344 (47 Mb) & 640 (180 Mb)  \\\hline
    \textbf{GIN}     & 817 & 66,326 & 81,144 & 365 & 10 \\
    \textbf{GIN+VN}     & 817  & 68,204 & 82,498 & 367 & 10 \\
    \textbf{GCN}     & 424 & 173,899 & 375,882 & 203 & 0 \\
    \textbf{PNA*}     & 50 & 40,951 & 34,533 & 233 & 144 \\
    \textbf{GAT}     & 341 & 80,545 & 82,829 & 484 & 0 \\
    \textbf{DGN}     & 1,042 & 73,735 & 93,579 & 523 & 0 \\
         
    \bottomrule
    \end{tabular}
    \caption{Resource utilization on Xilinx Alveo U50 FPGA. The clock frequency is 300 MHz. Results for the PNA* model are currently estimates from the Vitis HLS tool.}
    \label{tab:resource}
    \vspace{-10pt}
\end{table}}

\newcommand{\tabresourcelge}{%
\begin{table}
    \centering
    \footnotesize
    \begin{tabular}{c|c|c|c|c|c}
    \toprule
    \textbf{Dataset}    & \textbf{Nodes} & \textbf{Edges} & \textbf{Feat.\ Dim.} & \textbf{LUT} & \textbf{FF} \\ 
    \midrule
    \textbf{Cora} & 2,708 & 10,556 & 1,433 & 111,456 & 110,508 \\
    \textbf{CiteSeer} & 3,327 & 9,104 & 3,703 & 116,442 & 109,765 \\
    \textbf{PubMed} & 19,717 & 88,648 & 500 & 119,329 & 100,699 \\
    \bottomrule
    \end{tabular}
    \caption{The three datasets used to test DGN with Large Graph Extension, along with their corresponding resource utilization. GenGNN with the Large Graph Extension utilizes 1,344 DSPs, 494 BRAMs, and 0 URAMs for all three datasets.}
    \label{tab:resource_lge}
    \vspace{-10pt}
\end{table}}

\newcommand{\figpipelineeval}{%
\begin{figure*}
    \centering
    \includegraphics[width=0.99\textwidth]{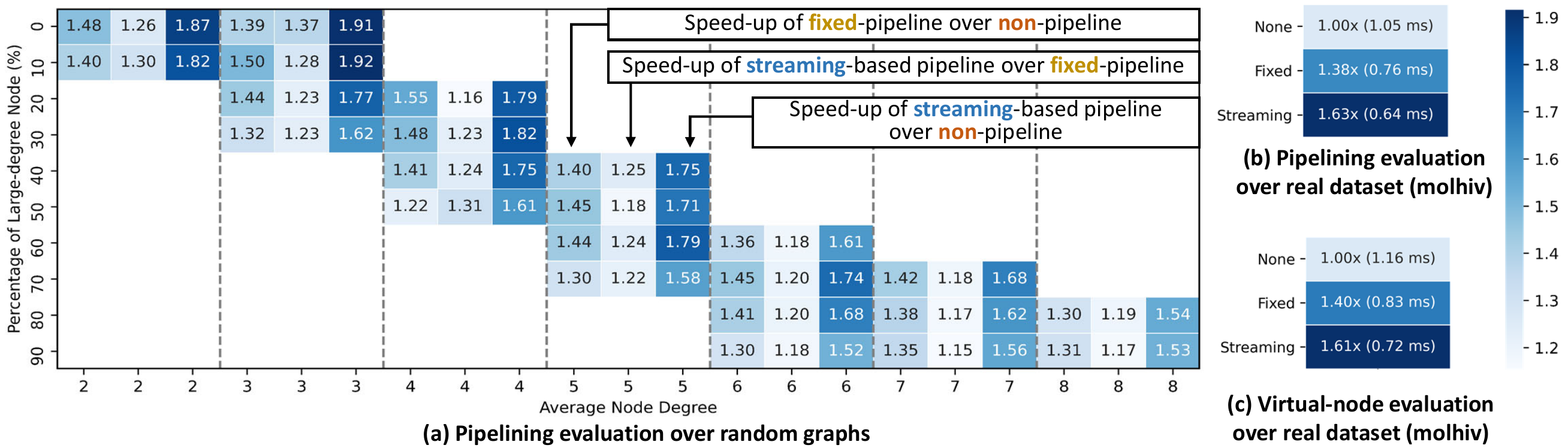}
    \vspace{-6pt}
    \caption{Speed-up of NE/MP pipelining techniques: fixed- and streaming-based pipeline over non-pipeline. Fixed-pipeline is up to $1.5\times$ faster than non-pipeline; streaming-based  is up to $1.37\times$ faster than fixed and $1.92\times$ faster than non-pipeline.}
    \label{fig:pipeline-eval}
\end{figure*}}

\newcommand{\figginmlp}{%
\begin{figure}
    \centering
    \includegraphics[width=0.45\textwidth]{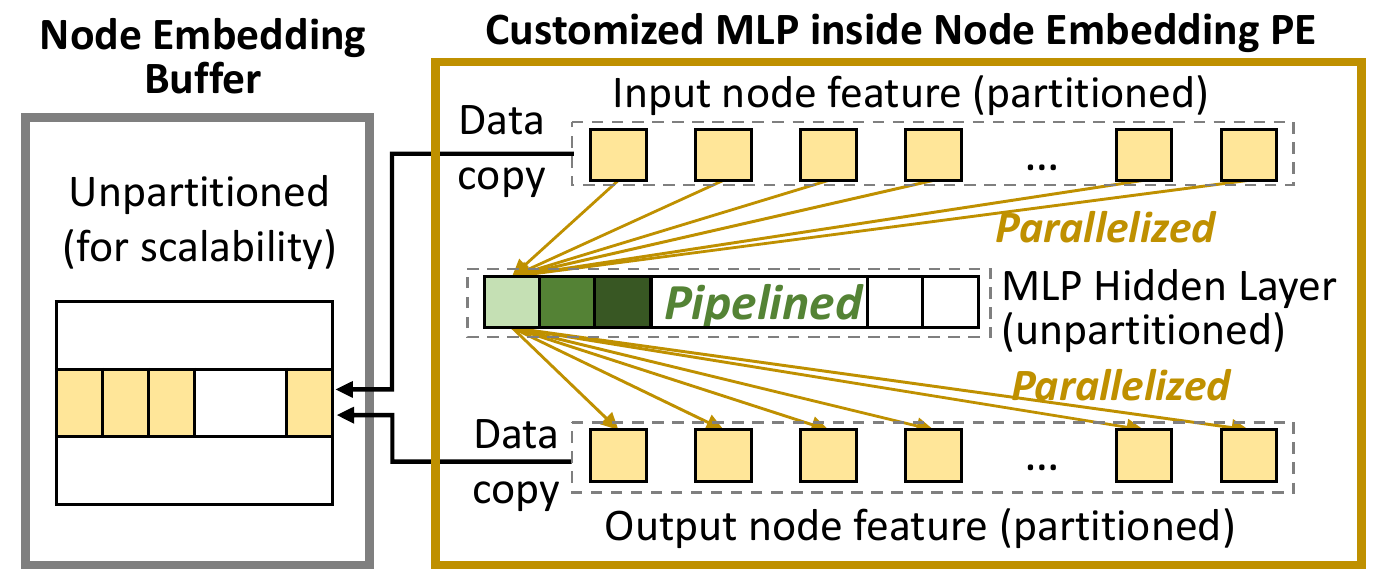}
    \caption{Customized MLP for GIN for node embedding. %
    }
    \label{fig:GIN-MLP}
\end{figure}}

\newcommand{\tabgnnnotation}{%
\begin{table}
    \centering
    \footnotesize
    \begin{tabular}{c|c}
    \toprule
    Notation & Description \\ \midrule
     $\mathcal{G}=(A, X, E)$    & The input graph for the GNN \\
    $A \in \mathbb{R}^{N \times N} $     & Adjacency matrix of $\mathcal{G}$ ($N$: number of nodes)\\
    $x_i^l \in X\in \mathbb{R}^{N \times F} $ & Node embedding of node $i$ at layer $l$ \\
    $e_{i,j}^l \in E \in \mathbb{R}^{M \times D} $ & Edge embedding of edge ${i, j}$ at layer $l$\\
    $F, D$ & Dimension of node and edge embeddings \\
    $\mathcal{N}(i)$ & All neighbors of node $i$\\
    $\phi(\cdot)$ & Differentiable message transformation function\\
    $\mathcal{A}(\cdot)$ & Permutation-invariant aggregation function\\
    $\gamma(\cdot)$  & Differentiable node transformation function \\

    \bottomrule 
    \end{tabular}
    \caption{Notations used in GNN definition.}
    \label{tab:GNN-notation}
    \vspace{-10pt}
\end{table}}

\makeatletter
\renewcommand\@authornotes{%
  \footnotetext[1]{Equal contribution; ordered alphabetically by last name.}
  \footnotetext[4]{Work done during internship at the Georgia Institute of Technology.}}
\makeatother

\begin{document}
\title{GenGNN: A Generic FPGA Framework for Graph Neural Network Acceleration}

\author{Stefan Abi-Karam\footnotemark[1]\footnotemark[2], Yuqi He\footnotemark[1]\footnotemark[2], Rishov Sarkar\footnotemark[1]\footnotemark[2], Lakshmi Sathidevi\footnotemark[2], Zihang Qiao\footnotemark[3]\footnotemark[4], Cong Hao\footnotemark[2]}
\affiliation[obeypunctuation=true]{\footnotemark[2]Georgia Institute of Technology, School of Electrical and Computer Engineering, Atlanta, GA, \country{USA}}
\affiliation[obeypunctuation=true]{\footnotemark[3]Beijing University of Posts and Telecommunications, Beijing, \country{China}}
\affiliation[obeypunctuation=true]{\{\href{mailto:stefanabikaram@gatech.edu}{stefanabikaram}, \href{mailto:yhe374@gatech.edu}{yhe374}, \href{mailto:rishov.sarkar@gatech.edu}{rishov.sarkar}, \href{mailto:lsathidevi3@gatech.edu}{lsathidevi3}\}@gatech.edu, \href{mailto:2018211889@bupt.cn}{2018211889@bupt.cn}, \href{mailto:callie.hao@ece.gatech.edu}{callie.hao@ece.gatech.edu}\country{}}

\settopmatter{printacmref=false} %
\renewcommand\footnotetextcopyrightpermission[1]{} %
\pagestyle{plain} %

\begin{abstract}
Graph neural networks (GNNs) have recently exploded in popularity thanks to their broad applicability to ubiquitous graph-related problems such as quantum chemistry, drug discovery, and high energy physics.
However, meeting demand for novel GNN models and fast inference simultaneously is challenging because of the gap between the difficulty in developing efficient FPGA accelerators and the rapid pace of creation of new GNN models. %
Prior art focuses on the acceleration of specific classes of GNNs but lacks the generality to work across existing models or to extend to new and emerging GNN models.
In this work, we propose a generic GNN acceleration framework using High-Level Synthesis (HLS), named \textbf{GenGNN}, with two-fold goals.
First, we aim to deliver ultra-fast GNN inference without any graph pre-processing for real-time requirements.
Second, we aim to support a diverse set of GNN models with the extensibility to flexibly adapt to new models. %
The framework features an optimized message-passing structure applicable to all models, combined with a rich library of model-specific components. 
We verify our implementation on-board on the Xilinx Alveo U50 FPGA and observe a speed-up of up to 25$\times$ against CPU (6226R) baseline and 13$\times$ against GPU (A6000) baseline. 
Our HLS code will be open-source on GitHub upon acceptance.\footnote{\href{https://github.com/sharc-lab/GenGNN}{https://github.com/sharc-lab/GenGNN}}

\end{abstract}

\maketitle

\section{Introduction}

Graph neural networks (GNNs) have become a powerful tool for applying deep learning to solve tasks infilling graph structures. %
Learning tasks for graphs can be applied at node-level (e.g., presence of protein~\cite{szklarczyk2019string}), edge-level (e.g., drug-drug interactions~\cite{wishart2018drugbank}), and graph-level (e.g, molecular property prediction~\cite{wu2018moleculenet}). %
Representative applications of GNNs include analysis of social networks and citation networks, recommendation systems, traffic forecasting, LIDAR point cloud segmentation for autonomous driving, high energy particle physics, and molecular representations~\cite{atz2021geometric}.

Targeting various applications, there is a huge demand for GNN inference acceleration with diverse \textbf{requirements}.
For instance, point cloud segmentation and detection for autonomous driving~\cite{shi2020point} and high energy particle physics~\cite{shlomi2020graph} require \textit{real-time processing}; in a particular example, the collision data from particle collider are collected every $25ns$ and thus must be processed using GNNs within nanoseconds with raw input graphs~\cite{iiyama2021distance, cerminara2020distance}. 
For social network applications, the size of the graphs to be processed is usually extremely huge and the computation time and memory cost are significant; therefore, such applications are in demand for GNN accelerators for \textit{large-scale} graphs.
Consequently, hardware acceleration is critical to apply GNNs to these applications and address real-time or large-scale computation.

The \textbf{challenges} and the \textbf{limitations} of existing accelerators, however, are significant.
\underline{First}, GNN computation is both 
communication-intensive and computation-intensive, as also noted by previous literature~\cite{zhang2020hardware, Li2021meloppr, auten2020hardware}, involving massive irregular memory access for message passing and heavy computation for embedding transformation.
\underline{Second}, novel GNN models are rapidly emerging while the accelerator innovation is lagging behind. For instance, most state-of-the-art GNN accelerators are tailored for graph convolutional networks (GCNs)~\cite{geng2020awb, zhang2020hardware, zhang2021boostgcn}, which can be conveniently expressed as sparse matrix multiplications (SpMM). However, the majority of GNNs are not suitable for SpMM because of complicated operations such as edge embedding, attention, mixed neighborhood aggregation, etc. Therefore, \textit{generic, extensible, and flexible acceleration frameworks} are required to rapidly adapt to evolving GNN models.
\underline{Third}, some accelerators adopt graph preprocessing to employ data locality~\cite{zhang2020hardware, jia2020redundancy, wang2020gnnadvisor, chen2021rubik, yan2020hygcn}, while some apply graph partitioning relying on the property of a fixed input graph (e.g., by analyzing the adjacency matrix sparsity~\cite{zhang2021boostgcn}).
Such preprocessing or graph-specific techniques \textit{are not feasible} for real-time applications with millions of input graphs with varied structures.

\tabfpgasotacompare
Motivated by the emerging requirements and existing limitations, we propose a generic and flexible framework on FPGA for GNN acceleration, named \textbf{GenGNN}, which supports a wide range of prevailing GNNs and is extensible for new models.
Highlighting the features of \textbf{GenGNN} in Table~\ref{tab:fpga-sota-compare}, we summarize our contributions as follows:
\begin{itemize}[leftmargin=*]
\tabgnnlist
    \item {GenGNN is the \textit{first generic} GNN acceleration framework that can process a large variety of GNNs with on-board implementation. Table~\ref{tab:GNN-list} summarizes the currently supported GNN models; each is a representative of a large GNN family. The framework is also dataset and graph structure agnostic, i.e., does not rely on the analysis for a specific input graph.}
    \item {GenGNN is also the first to target \textit{real-time} applications with \textit{zero preprocessing} and partitioning, where the graphs are streamed in consecutively.}
    \item {GenGNN is developed using High-Level Synthesis (HLS) for ease-of-use with modularized processing elements (PEs) for GNN components for extensibility. We will open-source the framework with an on-board executable for each GNN model upon acceptance.}
    \item {GenGNN is composed of a novel and highly optimized \textit{message passing} architecture, which accommodates the majority of state-of-the-art GNNs. It also includes a rich library of model-specific components, where new GNN components can be developed and plugged in seamlessly.}
    \item {We extensively conduct experiments on two popular datasets with more than 45k graphs and three widely used large graphs. Our measurements are \textit{on-board} execution with guaranteed \textit{end-to-end correctness} by cross-checking with PyTorch implementation. \textbf{GenGNN} on Xilinx Alveo U50 FPGA achieves a speed-up of up to $25\times$ against CPU (6226R) baseline and $13\times$ against GPU (A6000) baseline.
}
\end{itemize}

\section{Related Work}

GNN acceleration is attracting intensive attention in the research community. Recent works are summarized by a survey~\cite{abadal2021computing}, including both ASIC and FPGA accelerators. %

The majority of the accelerators are targeting ASICs via simulation. For instance, HyGCN~\cite{yan2020hygcn} is one of the earliest which introduces a hybrid architecture for GCN acceleration. 
EnGN~\cite{liang2021engn} uses PEs connected in a ring and performs aggregations using a technique called Ring-Edge Reduce, while GRIP~\cite{kiningham2020grip} uses the GReTA abstraction \cite{kiningham2020greta} to enable acceleration of any GNN variant.
AWB-GCN~\cite{geng2020awb} aims to combat workload imbalance in graph processing, while
GCNAX \cite{li2021gcnax} addresses the shortcomings of resource underutilization and excessive data movement using a flexible dataflow.%

On the other hand, FPGA-based accelerators primarily focus on GCN acceleration.
Zhang \textit{et al.} \cite{zhang2020hardware} combine software preprocessing with hardware utilizing both node-level and feature-level parallelism,
while BoostGCN~\cite{zhang2021boostgcn} specifically optimizes GCN via sparsity analysis and graph partitioning.
Auten \textit{et al.} \cite{auten2020hardware} propose an architecture that uses general-purpose CPUs connected by a network-on-chip to accelerate various GNNs. 
Rubik~\cite{chen2021rubik} and
GraphACT \cite{zeng2020graphact} aim to accelerate GCN training using ASIC and FPGA, respectively.

\subsubsection*{Limitations}
The computation pattern of GCN can be abstracted as a series of sparse and dense matrix multiplications so that most accelerators focus on optimizing the SpMM kernels.
However, advanced GNNs such as PNA, GAT, and GIN require more complex compute patterns than SpMM. These models use aggregators besides sum aggregation and require materialization of neighboring nodes to compute edge embeddings,
which must be done explicitly on a per-node basis.
In addition, many of these accelerators also adopt off-chip preprocessing such as graph partitioning, which in real-time applications is not possible.

In the following, we introduce the overall architecture of \textbf{GenGNN} in Section~\ref{sec:generic-gengnn}, and discuss the rich model-specific libraries in Section~\ref{sec:model-specific}. Performance is presented in Section~\ref{sec:results}.
\section{Generic Architecture of GenGNN}
\label{sec:generic-gengnn}

\subsection{GenGNN Framework Features}

\subsubsection*{GenGNN Features}
Our goal is to provide a generic and real-time FPGA acceleration framework for a wide range of GNNs, with great flexibility to support emerging GNNs with minimum modification in a plug-in manner.
Table~\ref{tab:fpga-sota-compare} summarizes our key features compared with state-of-the-art FPGA accelerators. \circled{1} \textbf{Generic.} While most existing FPGA works are limited to GCN by formulating GNN computation as SpMM, \textbf{GenGNN} supports a wide variety of GNN types by constructing a \textit{message passing} architecture to which most GNNs belong. It can also accept arbitrary dataset and graph structures.
\circled{2} \textbf{Real-time.} In practical applications such as point clouds in autonomous driving and particle graphs in high energy physics, the graphs are constructed and must be processed in real-time, leaving no time for preprocessing, partitioning, or adjacency matrix analysis. Different from accelerators that rely on a preprocessing/partitioning optimization, \textbf{GenGNN} directly processes raw graphs consecutively, targeting real-time inference for a large number of graphs.
\circled{3} \textbf{Open-source.} We will open-source our framework written in HLS to encourage more GNNs to be built on top of \textbf{GenGNN} to promote a GNN acceleration ecosystem.

\subsubsection*{Supported GNNs}
Table~\ref{tab:GNN-list} summarizes currently supported GNNs, each being representative of a family of GNNs. 
Graph convolutional network (GCN)~\cite{kipf2016semi} represents the ones can be formulated as sparse matrix-matrix multiplications (SpMM); simplified GCN~\cite{wu2019simplifying} also falls into this category. 
Graph isomorphism network (GIN)~\cite{xu2018powerful} represents advanced GNNs with higher representation power, including edge embeddings and transformations where SpMM \textit{does not} apply; GraphSage~\cite{hamilton2017inductive} falls into this category.
Principal neighborhood aggregation (PNA)~\cite{corso2020principal} represents a popular GNN family that uses multiple arbitrary aggregation methods simultaneously.
Graph attention network (GAT)~\cite{velivckovic2017graph} represents a GNN family with multi-head self-attention and possibly with edge embeddings.
Directional graph network (DGN)~\cite{beani2021directional} is a state-of-the-art GNN with directional flow at nodes with guided aggregation.
GNN with virtual node (VN)~\cite{gilmer2017neural} is a widely used GNN technique using virtual nodes connected to all other nodes. 
In addition, GenGNN supports both node-level and graph-level tasks (easily extended to edge-level tasks), and is scalable to both small and large graphs.

\figsparseformats
\subsection{Graph Data Representation}
\label{sec:sparse-format}

The input graphs for GNNs typically have sparse adjacency matrices, which can be stored in different formats.

\subsubsection*{COO}
In \textbf{COO}rdinate format, edges are stored in an arbitrarily-ordered list, where each list entry consists of the source node, the destination node, and the data associated with the edge. COO format is convenient for graph producers and is usually the raw graph format in real-time applications, but can be less performant for graph consumers.

\subsubsection*{CSR}
In \textbf{C}ompressed \textbf{S}parse \textbf{R}ow format, adjacency information is stored in three arrays. The first is the degree table, where entries denote the out-degrees of each node. The second is the neighbor table, containing the row-major concatenation of the out-neighbors. Fig.~\ref{fig:sparse_formats} presents an example. Each consecutive slice (marked the same color) lists destination nodes coming from each source node. The third array (not shown) stores edge data. 
CSR format is beneficial when processing all edges with the same source node consecutively.

\subsubsection*{CSC}
\textbf{C}ompressed \textbf{S}parse \textbf{C}olumn format is similar to CSR but using column-major order. In CSC format, the degree table stores the in-degree of each node, and the neighbor table is a column-major concatenation of the in-neighbors of each node. CSC format is beneficial when processing all edges with the same destination node consecutively.

Since the raw input graphs are in COO format, we develop an on-chip converter to transform into CSR or CSC as needed. The converter runs once when the graph is streamed into the FPGA and is reused for all the GNN layers.

\figgnnoverall
\figarchoverall

\subsection{Message Passing Mechanism}
\label{sec:message-passing-mech}

Most prevailing GNN architectures follow the message passing mechanism~\cite{wang2020gcn, gao2019graph, wu2019simplifying, kipf2016semi, xu2018powerful, hamilton2017inductive, corso2020principal, velivckovic2017graph, beani2021directional, gilmer2017neural}.
\tabgnnnotation
With notation defined as in Table~\ref{tab:GNN-notation}, the general computation of a message passing GNN can be expressed as:
\begin{equation*}
x_i^{l+1} = \gamma(x_i^l, \mathcal{A}_{j\in \mathcal{N}(i)}(\phi(x_i^l, x_j,l, e_{i,j}^l))    
\end{equation*}

Fig.~\ref{fig:gnn-overall} demonstrates the message passing procedure for a single node $n_1$ at layer $l$, which will be repeatedly applied for all nodes and for several layers.
Highlighted at the bottom of the figure, there are two major steps for each node in each layer of the GNN: \textit{message passing (MP)} and \textit{node embedding (NE)}.
Message passing can be divided into ``gather'' and ``scatter'' phases, where ``gather'' corresponds to feature aggregation, and ``scatter'' corresponds to message transformation and passing. Node embedding involves node transformation and update.

\subsubsection*{Message Passing (Gather)}
In the ``gather'' phase, a.k.a., aggregation, of a certain node $n_1$, the messages from its neighboring nodes obtained in the previous layer are retrieved from a message buffer. The messages are then aggregated in a permutation invariant manner, denoted by $\mathcal{A}(\cdot)$ (e.g.,  sum, max, mean, std.\ dev.). In advanced GNNs such as PNA, multiple aggregators are used with learnable weights and scaled according to the degree of the target node. The aggregated message is denoted by $m_{1}^{l}$.

\subsubsection*{Node Transformation}
After aggregation, $m_{1}^{l}$ is processed together with node $n_1$'s current node embedding, denoted by $x_1^l$, via a node embedding function, $\gamma(\cdot)$.
$\gamma(\cdot)$ applies a node transformation using $m_{1}^{l}$ and $x_1^l$, such as
identity function, fully-connected layer, weighted sum of $m_{1}^{l}$ and $x_{1}^{l}$, or an MLP applied to weighted sum or concatenation of $m_1^l$ and $x_1^l$. 
After the transformation, $\gamma(\cdot)$ produces a new node embedding of $n_1$, denoted by $x_{1}^{l+1}$, and applies the update.

\subsubsection*{Message Passing (Scatter)}
After node transformation is the ``scatter'' phase of message passing. The new node embedding $x_{1}^{l+1}$ will be transformed by a message transformation function, $\phi(\cdot)$, usually together with an edge embedding $e_{src, dest}^{l+1}$, to generate the node's outgoing message.
The message will be dispatched to all its neighbors, which will eventually be collected by the ``gather'' stage of the next layer.%

A complete GNN model may consist of multiple layers, each with message passing and node embedding steps. %
For graph-level tasks, a global pooling layer
is needed, possibly followed by MLP layers for final prediction.

\subsection{High Level Hardware Framework}
\label{sec:high-level-framework}

\subsubsection*{PEs and Buffers}
We develop GenGNN following the message passing style.
Fig.~\ref{fig:arch-overall} illustrates the generic architecture.
It has two main processing elements (PEs): node embedding (yellow block) and message passing (blue block), corresponding to the two steps described in Section~\ref{sec:message-passing-mech}.
It has three data storage buffers: one node embedding buffer and two message buffers, all of size $O(N)$ where $N$ is the number of nodes allowed on-chip. We temporarily assume that the whole graph can be stored on-chip and will discuss the extension to larger graphs that cannot fit in Section~\ref{sec:large-graph}.
The two message buffers act alternately across layers. For instance, message buffer 1 is read-only during layer 1 while message buffer 2 is being updated; during layer 2, message buffer 2 becomes read-only and buffer 1 updates.

\subsubsection*{Execution Flow}
Fig.~\ref{fig:arch-overall} illustrates the execution flow of one GNN layer. For multiple layers, the same resources and dataflow will be reused.
Within one layer,
the \textbf{node embedding PE} applies node transformation and update, e.g., MLP, activation, and self-attention. This is the main component that distinguishes different GNN models.
Then, the \textbf{message passing PE} performs the subsequent scatter operation.
Consider the graph in Fig.~\ref{fig:gnn-overall} as an example. Once node $n_1$'s embedding is updated, the MP PE retrieves its neighbors, i.e., $n_2$, $n_3$, and $n_4$, computes the messages together with edge embeddings $e_{1,2}$, $e_{1,3}$, and $e_{1,4}$, and dispatches the messages.
The receivers will instantly update their partially aggregated message in the message buffer.
In this way, the scatter and gather phases can be merged, because the aggregation function is permutation invariant and the order of aggregation does not matter.
Such a merged fashion has two merits. First, it reduces the overall process latency by fusing two stages into one. Second, it reduces memory cost from $O(E)$ to $O(N)$ where $E$ is the number of edges and is typically much larger than $N$.
This approach requires that the graph data is stored in its CSR format.

Note that an equivalent procedure is to first perform gather, i.e., aggregation, via incoming edges and then perform node transformation, in which case no scatter is needed. This approach requires two node embedding and one message buffer, all of size $O(N)$, and graph data stored in CSC format.

The NE and MP stages are connected by a FIFO (first-in first-out) queue, which enables pipelined execution. Details will be discussed in Section~\ref{sec:node-edge-pipelining}.

\subsection{Node Embedding/Message Passing Pipeline}
\label{sec:node-edge-pipelining}

\figpipeliningillustration

As demonstrated in Fig.~\ref{fig:gnn-overall}, the two major steps, node embedding (NE) and message passing (MP), are dependent for a particular node but are independent across nodes and edges.
This presents an opportunity to pipeline NE and MP to largely reduce processing latency.

Fig.~\ref{fig:pipelining-illustration} explains three strategies for NE and MP processing.
\circled{1}~\textbf{Non-pipelining.}
In Fig.~\ref{fig:pipelining-illustration}(a), NE and MP are not pipelined, which apparently suffers from a huge waste of idle cycles.
\circled{2}~\textbf{Fixed pipelining.}
In Fig.~\ref{fig:pipelining-illustration}(b), NE and MP are pipelined in a fixed manner: NE for the second node is pipelined with MP for the first node, etc. This achieves some latency reduction but suffers from imbalanced node degree. Specifically, if some nodes have larger degrees and their MP latency is longer than NE, while others have shorter MP latency, there still will be idle cycles.
\circled{3}~\textbf{Streaming-based pipelining.}
In Fig.~\ref{fig:pipelining-illustration}(c), NE and MP are pipelined flexibly using a node queue:
as soon as a node finishes its NE, i.e., is ready for message passing, its embeddings are pushed into the queue; meanwhile, the MP engine will read from the queue and fetch the node embeddings for message passing. This scheme can be implemented using a streaming-based FIFO (first-in first-out) memory queue. This approach can greatly reduce the idle cycles and minimize the resource. %

We acknowledge that our NE/MP pipeline shares a similar idea with the task scheduling in BoostGCN~\cite{zhang2021boostgcn}. The differences are: 1) BoostGCN first sorts the vertices based on their degrees to determine an execution order on CPU, whereas we process on-the-fly in FPGA adaptively; 2) BoostGCN uses a buffer in external memory, whereas we build an on-chip FIFO to queue the nodes that are ready for message passing.

\section{Model-Specific Components}
\label{sec:model-specific}

\subsection{Graph Isomorphism Network}

\figginmlp

GIN is representative of the GNN family whose message passing involves edge embeddings, and whose node transformation is computation intensive using MLPs.
Each node's outgoing message is a weighted sum of its own node embedding and the outgoing edge embedding. Therefore, GIN adopts the CSR format converted from COO on-chip. Its message passing is within the framework using a customized message transformation function $\phi(x, m)=x^l + \epsilon^l \cdot m^l$.

For the computation-intensive MLP, we develop a customized MLP PE inside the node embedding PE, as shown in Fig.~\ref{fig:GIN-MLP}. For scalability and generality, the principle is \textit{not} to apply optimizations to the node embedding and message buffers, but rather to allocate local buffers upon necessity.
Therefore, the MLP PE has two local buffers fully-partitioned for inputs/outputs of one node, copied from/to the global node embedding buffer. Using ping-pong buffers, the data copy latency is overlapped with the MLP computation.
We parallelize the multiplications at the partitioned input and output buffers and pipeline the execution along the MLP hidden layer elements.
The designed MLP PE can be reused for other GNNs since MLP is a very common GNN operation.

\subsection{Graph Attention Network}

GAT is representative for its multi-head self-attention: each node's incoming messages are first weighted before aggregation, where the attention coefficients are computed using the node and its neighbor's embeddings. This exposes additional computation complexity, but fortunately, GAT is still fully compatible with GenGNN.
Similar to GIN, GAT also needs a customized message transformation function, where $\phi(x, m)=x_i^l + \sigma_{i,j}^l \cdot m_j^l$, where $\sigma_{i,j}$ is the attention coefficient from node $j$ to node $i$, computed using an attention function $\sigma_{i,j} = A(x_i, x_j)$ such as weighted sum or MLP.

The complexity of multi-head attention is proportional to the number of heads. Therefore, we parallelize along the head dimension in order to attain a speed-up while keeping the original node embedding and message buffers intact.

\subsection{Principled Neighbor Aggregation}

PNA uses multiple neighbor aggregations to increase the distinguishing power, where the node embedding is computed following~\cite{corso2020principal}:
$ x_i^{l+1} =\text{relu}(\text{linear}(\bigoplus_{j\in \mathcal{N}(i)} (x_j^l)))$, and
\begin{equation*}
    \bigoplus = 
        \begin{bmatrix}
            1\\[\medskipamount]
            \log(D_i+1) / \widetilde{D}\\
            \widetilde{D} / \log(D_i+1)\\
        \end{bmatrix}
    \otimes
        \renewcommand{\arraystretch}{0.8}
        \begin{bmatrix}
            \mu \\
            \sigma \\
            \max \\
            \min
        \end{bmatrix}
\end{equation*}
where $D_i$ is the degree of $x_i$ and $\widetilde{D}$ is the average node degree seen in training data.

PNA falls into the message passing framework with a difference at the aggregation function.
It has four aggregators including min, max, mean, and standard deviation; each aggregator stores its results into a separate buffer. All the scaling values are then computed and multiplied with the four aggregation values and written into a global buffer. Lastly, a pipelined linear-ReLU kernel is applied to compute the new node embedding, which reuses the MLP in the GIN model. %
Skip connections are added after each PNA layer computation to copy and accumulate embeddings from the output of the previous layer.

\subsection{Directional Graph Networks}
\label{sec:dgn}

DGN~\cite{beani2021directional} uses vector fields in a graph to define directional flows at each node that can be used for graph convolutions using anisotropic kernels. It uses eigenvectors of the graph Laplacian matrix to define directional aggregation matrices used in the ``gather'' phase of message passing.

Similar to its baseline PyTorch implementation, DGN accepts the precomputed Laplacian eigenvectors as a parameter and uses them to compute the relevant directional aggregation matrices on-the-fly during message passing.
DGN uses two aggregations: the mean, and the directional derivative aggregation along the first eigenvector as $Y^l = \text{concat}\{ D^{-1}AX^l, |B_{dx}^lX^l| \}$
where ${Y}^{l}$ is the aggregated messages, the ${X}^{(t)}$ is the node embeddings, ${D}$ is the degree matrix, ${A}$ is the adjacency matrix, and ${B}_{dx}^1$ is the directional derivative matrix along first eigenvector.
GenGNN is trivially extensible to other types of DGN aggregations, including directional smoothing ${B}_{av}$.

DGN's node transformation uses an MLP with skip connections similar to PNA.
The aggregation components run concurrently, as do the node transformations using each of the concatenated messages, enabling a total time complexity of \(O(E + N)\) for each DGN layer.

\subsection{Virtual Node}

\figvirtualnode

Our proposed streaming-based pipelining for node/edge processing (Section~\ref{sec:node-edge-pipelining}) is especially beneficial for models with virtual nodes.
A virtual node~\cite{gilmer2017neural} is an artificial node connected to all other nodes in the graph. The virtual node provides a shortcut for message passing between node pairs, which is demonstrated to be effective in many GNN models~\cite{ishiguro2019graph, xue2021node, pham2017graph}.
As illustrated in Fig.~\ref{fig:virtual-node} left, a virtual node is \textit{busy} with connections to all nodes, which results in highly unbalanced workload especially for large graphs and thus require special processing. In some models, there can be multiple virtual nodes~\cite{xue2021node} which escalates the model complexity.

Fortunately, benefiting from our proposed streaming-based pipelining for node/edge processing, the imbalance introduced by virtual nodes can be easily resolved without changing the framework.
As shown in Fig.~\ref{fig:virtual-node} right top, in fixed-pipeline or no-pipeline architectures, the process for the virtual node itself will take much longer time than others, resulting in a large waste.
In contrast, as shown in Fig.~\ref{fig:virtual-node} right bottom, in our proposed streaming-based architecture, processing of the virtual node can be fully overlapped with the node embedding computation for other nodes, with zero waste, as long as it is processed early enough (depending on the node ID numbering and processing order, which is adjustable).
Effectiveness will be evaluated in Section~\ref{sec:streaming-eval}.

\subsection{Large Graph Extension}
\label{sec:large-graph}

For large graphs that do not fit on-chip, our message passing mechanism is still applicable, i.e.,  the streaming-based pipeline working with two PEs, node embedding and message passing.
However, additional strategic optimizations are necessary to hide memory access latency.

\subsubsection*{Prefetching}
Synthesis of loop-carried dependences involving DRAM access can result in substantial overhead and is usually hard to pipeline.
Since the graph is too big, its neighbor list stored in CSR format must be fetched from DRAM, which disrupts the pipeline inside message passing PE and introduces extra latency.
To avoid this overhead, we develop a prefetcher, which fetches the degrees of consecutive nodes from DRAM into an on-chip FIFO buffer. The message passing PE loads each consecutive node's degree as needed, triggering the prefetcher to refill the buffer. In this way, the latency of fetching from the off-chip degree table is completely hidden, and the message passing PE behaves in the same way as for small graphs.

\subsubsection*{Packed Data Transfers}
For large graphs, the node embedding and message buffers are stored off-chip, while only the streaming FIFO and the prefetching results are stored on-chip. 
To alleviate the data transfer overhead and to saturate the AXI bus bandwidth, the off-chip data transfer must fully utilize the bus width.
For example, transferring one 16-bit array element per clock cycle is a waste for a 64-bit bus. Therefore, we apply packed data transfer by typecasting off-chip array pointers to pointer types of a desired size, to transfer larger numbers of bits from and to DRAM each clock cycle.
Given four 64-bit AXI buses, we pack 8 16-bit values and parallelize the fetching in one cycle, with an unpacking module reads 8 values in parallel.
A similar technique is used to reduce DRAM writeback latency.

\section{Experimental Results}
\label{sec:results}

\subsection{FPGA Implementation}
\tabresource
We implement \textbf{GenGNN} on Xilinx FPGA Alveo U50 accelerator card using Vitis HLS and Vivado.
The available resources of U50 is shown in Table~\ref{tab:resource}, and the FPGA logic runs at 300 MHz clock frequency.
The graphs are streamed into the FPGA in their raw edge-list format (i.e., COO) consecutively with zero CPU intervention.

As listed in Table~\ref{tab:GNN-list}, we implement six GNN models, each being representative to a family of GNNs.
Notably, each GNN model has a PyTorch version of implementation, to which we cross-check our on-board implementation and guarantee that our end-to-end execution is correct.
For GCN, GIN, and GIN-VN, the number of layers is 5 and the node embedding dimension is 100, as specified in the PyTorch model~\cite{ogb-models}. All of these models also use global average pooling, and an output head with a single linear layer.
For PNA, we use 4 layers with an node embedding dimension of 80, global average pooling, and an MLP-ReLU head with sizes $(40,20,1)$.
For DGN, we use 4 layers and a node embedding dimension of 100, global average pooling, and an MLP-ReLU head with sizes $(50,25,1)$.
For GAT, we use 5 layers with 4 heads and 16 features per layer, global average pooling, and an output head with a single linear layer.
\tabresourcelge
The resource utilization of all models are reported in Table~\ref{tab:resource},
and the resource utilization for Large Graph Extension is reported in Table~\ref{tab:resource_lge}.

We also would like to point out that we do not over-optimize the code, since this work is to demonstrate the architectural advantages of our framework. We deliberately omit some optimizations, for example, conservatively using 32-bit fixed point quantization (Large Graph Extension uses 16-bit); we also do not make optimizations over the parameters that may be unscalable (e.g., partitioning the dimension of maximum number of nodes).

The FPGA latency is measured \textit{end-to-end on-board}, obtained from the ``average execution time'' in the OpenCL summary report after the execution of all testing graphs.

\subsection{Dataset and Baseline}
\subsubsection*{Datasets}
We use three datasets to evaluate \textbf{GenGNN}.
For real-time processing evaluation, 
we adopt two molecular property prediction datasets, MolHIV and MolPCBA, from the Open Graph Benchmark~\cite{hu2020ogb}.
Both are \textit{graph-level} tasks.
MolHIV testing set has 4k graphs and MolPCBA testing set has 43k graphs.
For large-scale extension evaluation,
we adopt three prevailing classification datasets, CiteSeer, Cora~\cite{sen2008collective}, and PubMed~\cite{namata2012query}, where the graphs are much larger and cannot fit into on-chip memory.
All three are \textit{node-level} tasks.

\subsubsection*{Baseline}
We take CPU (Intel Xeon Gold 6226R) and GPU (NVIDIA RTX A6000) executions as the baseline.
We average five iterations of the time measured on CPU and GPU, where each model is implemented in PyTorch Geometric~\cite{pyg} with identical hyperparameters to their corresponding accelerator with batch size of 1.
Although we would like to compare with existing FPGA accelerators~\cite{zhang2020hardware, zhang2021boostgcn}, we cannot find the GNN details such as the number of layers and node embedding dimensions, and they only provide GCN results. More importantly, since we target real-time applications, we do not employ any preprocessing for the input graphs, while both~\cite{zhang2020hardware} and~\cite{zhang2021boostgcn} perform graph partitioning and~\cite{zhang2021boostgcn} conducts graph-specific optimizations. Therefore we find it difficult to make a fair comparison.

\subsection{End-to-end Evaluation}
\figlatencystats
We fully evaluate GenGNN using six GNN models by comparing with CPU and GPU baselines.
The results are depicted in Fig.~\ref{fig:latency_stats}.
The top figure uses the MolHIV dataset and the bottom figure uses the MolPCBA dataset.
It shows that for six GNN models, on MolHIV, GenGNN achieves 1.77--13.84$\times$ speed-up compared with CPU, and 2.05--25.96$\times$ speed-up compared with GPU;
on MolPCBA, GenGNN achieves 1.64--9.69$\times$ speed-up compared with CPU, and 1.92--17.66$\times$ speed-up compared with GPU.
It invariably demonstrates the effectiveness of GenGNN, especially given that we do not over-optimize our models but only exhibit the advantages of the framework itself.
Meanwhile, the most prominent speed-up is the DGN model (up to $25.9\times$), presumably because CPU and GPU are not specialized for the directional derivative aggregation (Section~\ref{sec:dgn}), which implies the necessity of customized FPGA accelerating components.

\figlargelatencystats
Next we evaluate the scalability of GenGNN to support large graphs that do not fit on-chip.
Table~\ref{tab:resource_lge} summarizes the graph size and node feature dimension (Feat. Dim.) for each benchmark.
Fig.~\ref{fig:lge_latency_stats} depicts the comparisons with CPU and GPU.
Compared with CPU, GenGNN achieves 1.49--1.95$\times$ speed-up; compared with GPU, it is 2.44$\times$ faster on Cora, 1.32$\times$ faster on CiteSeer, but 1.04$\times$ slower on PubMed.
Since the main focus of this work is not on large graphs, we will improve in future works.

\subsection{Streaming-based Pipelining Evaluation}
\label{sec:streaming-eval}

\figpipelineeval
Fig.~\ref{fig:pipeline-eval} demonstrates the effectiveness of the proposed streaming-based pipelining (Section~\ref{sec:node-edge-pipelining}) on both synthetic graphs and real benchmarks using the GIN model implementation.
First, we test 100k random graphs with various statistics in Fig.~\ref{fig:pipeline-eval}(a), including average node degree (x-axis) and the percentage of large-degree nodes (y-axis).
The x-axis is grouped every three columns by the same average node degree. Within each group, the first column is the speed-up of fixed-pipeline over non-pipeline (1.2$\sim$1.5$\times$); the second is the speed-up of streaming-based over fixed-pipeline (1.15$\sim$1.37$\times$); and the third is streaming-based over non-pipeline (1.53$\sim$1.92$\times$).

A general trend can be observed that when the average node degree is smaller and there are fewer high-degree nodes, the streaming pipeline is more beneficial. The reason is that the more imbalanced the latency of node embedding and message passing, the more beneficial streaming becomes in utilizing both PEs. When the node degree is large and the message passing dominates the latency (i.e., the node embedding latency is mostly shorter than the message passing), the streaming-based pipeline will degrade to fixed-pipeline.

We also verify its effectiveness on real benchmarks from the MolHIV dataset as well as with virtual nodes. Fig.~\ref{fig:pipeline-eval}(b) shows the speed-up of fixed and streaming pipelines, $1.38\times$ and $1.63\times$, respectively; Fig.~\ref{fig:pipeline-eval}(c) shows the speed-up with virtual nodes, $1.40\times$ and $1.61\times$, respectively.
This implies that the streaming-based pipeline is efficient in reducing processing latency. It also reduces memory cost since we set the queue depth to be 10 nodes.

\section{Conclusion}

In this work, we proposed \textbf{GenGNN}, the first generic and flexible FPGA accelerator framework for a wide range of GNNs.
Its noteworthy features include generality for future-proofing, real-time processing, and open-source.
GenGNN is composed of an optimized message-passing structure applicable to all models, combined with a rich library of model-specific components. 
On-board evaluation with guaranteed functionality exhibited invariant speed-up comparing with CPU and GPU baselines, as well as the effectiveness of our streaming-based pipeline architecture within message passing mechanism. 
We also demonstrated its scalability by extending to commonly used large graphs with speed-up.
Future work includes design automation, design space exploration, and large graph optimization for GenGNN.

\begin{acks}
The authors would like to thank Parima Mehta for her contribution to the development of the virtual node model.
The authors would also like to thank Dr. Pan Li for his insightful discussions.
\end{acks}

\bibliography{ref}
\bibliographystyle{ieeetr}

\end{document}